\documentclass{article}
\usepackage[preprint,nonatbib]{neurips_2019}
\usepackage[utf8]{inputenc} 
\usepackage[T1]{fontenc}    
\usepackage{hyperref}       
\usepackage{url}            
\usepackage{booktabs}       
\usepackage{amsfonts}       
\usepackage{nicefrac}       
\usepackage{microtype}      
\usepackage{graphicx}
\usepackage{bbm,bm}
\usepackage[ruled,vlined]{algorithm2e}

\usepackage{subfigure}
\usepackage{caption}
\usepackage{mathtools}
\usepackage{fancyvrb}
\usepackage{amsmath,amssymb,amsfonts,amsthm}
\DeclareRobustCommand{\[}{\begin{equation*}}
\DeclareRobustCommand{\]}{\end{equation*}}
\DeclareMathOperator*{\argmax}{argmax}
\DeclareMathOperator*{\argmin}{argmin}
\usepackage{color}
\usepackage[dvipsnames]{xcolor}
\usepackage{soul}
\usepackage{array}
\usepackage{cleveref}
\usepackage{floatrow}
\newfloatcommand{capbtabbox}{table}[][\FBwidth]
\usepackage[square,sort,comma,numbers]{natbib}

\newtheorem{prop}{Proposition}
\title{Defining Admissible Rewards for High-Confidence Policy Evaluation
}

\author{%
  Niranjani Prasad \\
  Dept. of Computer Science\\
  Princeton University\\
  \texttt{np6@princeton.edu} \\
   \And
   Barbara E Engelhardt \\
   Dept. of Computer Science \\
   Princeton University \\
   \texttt{bee@cs.princeton.edu} \\
   \And
   Finale Doshi-Velez \\
   Dept. of Computer Science \\
   Harvard University \\
   \texttt{finale@seas.harvard.edu} \\
}

\begin{document}

\maketitle

\begin{abstract}
A key impediment to reinforcement learning (RL) in real applications with limited, batch data is defining a reward function that reflects what we implicitly know about reasonable behaviour for a task and allows for robust off-policy evaluation.
In this work, we develop a method to identify an \emph{admissible} set of reward functions for policies that (a) do not diverge too far from past behaviour, and (b) can be evaluated with high confidence, given only a collection of past trajectories. Together, these ensure that we propose policies that we trust to be implemented in high-risk settings. We demonstrate our approach to reward design on synthetic domains as well as in a critical care context, for a reward that consolidates clinical objectives to learn a policy for weaning patients from mechanical ventilation.
\end{abstract}

\section{Introduction}
\label{introduction}

One fundamental challenge of reinforcement learning (RL) in practice is specifying the agent's reward. Reward functions implicitly define policy, and misspecified rewards can introduce severe, unexpected behaviour \citep{leike2017ai}. However, it can be difficult for domain experts to distill multiple (and often implicit) requisites for desired behaviour into a scalar feedback signal. 
Much work in reward design \citep{sorg2010internal, sorg2011optimal} or inference using inverse reinforcement learning \citep{abbeel2004apprenticeship, hadfield2017inverse, brown2017efficient} focuses on online, interactive settings in which the agent has access to human feedback \citep{loftin2014strategy, christiano2017deep} or to a simulator with which to evaluate policies and compare against human performance.  Here, we focus on reward design for \emph{batch} RL: we assume access only to a set of past trajectories collected from sub-optimal experts, with which to train our policies. This is common in many real-world scenarios where the risks of deploying an agent are high but logging current practice is relatively easy, as in healthcare, education, or finance \citep{doroudi2016sequence, atan2018learning}.

Batch RL is distinguished by two key preconditions when performing reward design. First, as we assume that data are expensive to acquire, we must ensure that policies found using the reward function can be \emph{evaluated} given existing data. Regardless of the true objectives of the designer, there exist fundamental limitations on reward functions that can be optimized and that also provide guarantees on performance. There have been a number of methods presented in the literature for safe, high-confidence policy improvement from batch data given some reward function, treating behaviour seen in the data as a baseline \citep{thomas2015highPI, ghavamzadeh2016safe, laroche2017safe, sarafian2018safe}. In this work, we turn this question around to ask: \emph{What is the class of reward functions for which high-confidence policy improvement is possible?}

Second, we typically assume that batch data are not random but produced by domain experts pursuing biased but reasonable policies. Thus if an expert-specified reward function results in behaviour that deviates substantially from past trajectories, we must ask whether that deviation was intentional or, as is more likely, simply because the designer omitted an important constraint, causing the agent to learn unintentional behaviour. This assumption can be formalized by treating the batch data as $\varepsilon$-optimal with respect to the true reward function, and searching for rewards that are consistent with this assumption \citep{huang2018learning}. Here, we extend these ideas to incorporate the uncertainty present when evaluating a policy in the batch setting, where trajectories from the estimated policy cannot be collected.

We can see that these two constraints are not equivalent. The extent of overlap in reward functions satisfying these criteria depends, for example, on the homogeneity of behaviour in the batch data: if consistency is measured with respect to average behaviour in the data, and agents deviate substantially from this average---e.g., across clinical care providers---then the space of policies that can be evaluated given the batch data may be larger than the policy space consistent with the average expert. 

In this paper, we combine these two conditions to construct tests for \emph{admissible} functions in reward design using available data. This yields a novel approach to the challenge of high-confidence policy evaluation given high variance importance sampling-based value estimates over extended decision horizons, typical of batch RL problems, and encourages safe, incremental policy improvement.
We illustrate our approach on several benchmark control tasks, and in reward design for a health care domain, namely, weaning a patient from a mechanical ventilator.

\section{Preliminaries and Notation}
\label{prelim}

A Markov decision process (MDP) is a tuple of the form $M = \{\mathcal{S}, \mathcal{A}, P_0, P, R, \gamma\}$, where $\mathcal{S}$ is the set of all possible states, and $\mathcal{A}$ are the available actions. $P_0(s)$ is the distribution over the initial state $s \in \mathcal{S}$; $P(s'|s,a)$ gives the probability of transition to $s'$ given current state $s$ and action $a \in \mathcal{A}$. $R(s,a)$ defines the reward for performing action $a$ in state $s$, and discount factor $\gamma \leq 1$ determines the relative importance of immediate and longer-term rewards. Our objective is to learn a policy $\pi^*: \mathcal{S}\rightarrow \mathcal{A}$ that maximizes expected cumulative discounted rewards---that is, $\pi^* = \argmax_\pi \mathbb{E}_{s \sim P_0} [V^\pi(s)|M]$---where the value function $V^\pi(s)$ is defined as:
\vspace{-1mm}
\begin{equation}
V^\pi = \mathbb{E}_{P_0,P,\pi}\left[\sum\limits_{t=0}^\infty \gamma^t R(s_t, a_t)  \right].
\end{equation}
\vspace{-4mm}

In batch RL, we have a collection of trajectories of the form $h= \{s_0,a_0,r_0,\dots,s_T,a_T,r_T\}$.  We do not have access to transition function $P$ or the initial state distribution $P_0$.
Without loss of generality, we express the reward as a linear combination of some arbitrary function of the state features: 
$R(s) = w^T\phi(s)$, where $\phi \in \mathbb{R}^k $ is a vector function of state features, and $||w||_1 = 1 $ (for invariance to scaling factors \citep{brown2017efficient}).  The value $V^\pi$ of a policy $\pi$ with reward weight $w$ can be written as:
\vspace{-1mm}
\begin{align}
V^\pi &= \mathbb{E}_{P_0,P,\pi} \left[\sum_{t=0}^{\infty}\gamma^t w^T\phi(\cdot) \right] \label{eqn:linear-reward}
= w^T\mu^\pi 
\text{, where }
\mu^\pi = \mathbb{E}_{P_0,P,\pi} \left[ \sum_{t=0}^{\infty}\gamma^t \phi(\cdot) \right]. 
\end{align}
where the vector $\mu^\pi$ denotes the \emph{feature expectations} \citep{abbeel2004apprenticeship} of policy $\pi$, that is, the total expected discounted time an agent spends in each feature state. Thus, $\mu^\pi$ provides a representation of state dynamics of a policy that is decoupled from the reward function.
To quantify confidence in the estimated value $V^\pi$ of policy $\pi$, we adapt the empirical Bernstein concentration inequality \citep{maurer2009empirical} to get a probabilistic lower bound $V_{lb}$ on the estimated value~\citep{thomas2015highPE}: consider a set of trajectories  $h_n,\, n \in 1...N$ and let $\hat{V}_n$ be the value estimate for trajectory $n$. Then, with probability at least $1-\delta$:
\vspace{-1mm}
{\small{\begin{equation}
   {V_{lb}} = \dfrac{1}{N}\sum\limits_{n=1}^N\hat{V}_n - \dfrac{1}{N}\sqrt{\dfrac{\ln(\frac{2}{\delta})}{N-1}  \smashoperator[r]{\sum\limits_{n,n'=1}^N}(\hat{V}_n-\hat{V}_{n'})^2} - \dfrac{7b\ln(\frac{2}{\delta})}{3(N-1)}, 
    \label{eqn:thomas-bound}
\end{equation}}}
where $b$ is the maximum achievable value of $V(\pi)$.

\section{Admissible Reward Sets}
\label{ars}

We now turn to our task of identifying admissible reward sets -- that is, defining the space of reward functions that yield policies that are \emph{consistent} in feature expectations with available observational data, as well as possible to \emph{evaluate} off-policy for high-confidence performance lower bounds. In Sections~\ref{crp} and~\ref{erp}, we define two sets of weights $\mathcal{P_C}$ and $\mathcal{P_E}$ to be the consistent and evaluable sets, respectively, show that they are convex, and define their intersection $\mathcal{P_C} \cap \mathcal{P_E}$ as the set of \emph{admissible} reward weights. In Sections~\ref{qarp}, we describe how to test whether a given reward lies in the intersection of these polytopes, and, if not, how to find the closest points within this space of admissible reward functions given some initial reward proposed by the designer of the RL agent.

\subsection{Consistent Reward Polytope}\label{crp}
Given near-optimal expert demonstrations, the \emph{consistent reward polytope} \citep{huang2018learning} is the set $\mathcal{P_C}$ of weights $w \in [-1, 1]^k$, for reward $R = w^T\phi(s)$, satisfying the assumption that the demonstrations achieve $\varepsilon$-optimal performance with respect to the ``true" reward.
We denote the behaviour policy of experts as $\pi_b$ with policy feature expectations $\mu_b$, where $V(\pi_b) = w^T\mu_b$. The set $\mathcal{P_C}$ is then all $w$ such that $ w^T\mu \leq w^T\mu_b + \varepsilon , \forall \mu \in \mathcal{P_F}$, where $\mathcal{P_F}$ is the space of all possible policy feature representations.
It has been shown that the set $\mathcal{P_C}$ is convex, given an exact MDP solver \citep{huang2018learning}.

Translating this to batch RL with a fixed set of \emph{sub-optimal} trajectories requires adaptations to both the constraints and the computation. First, we choose to constrain the relative rather than absolute difference in observed behaviour values and learnt policy, in order to handle high variance in the magnitudes of estimated values, and we make it symmetric so the value of the learnt policy can deviate equally above or below the value of the observed behaviour. This reflects the use of this constraint to place metaphorical guardrails on the deviation of the behaviour of the learnt policy from the policy in the batch trajectories, rather than to impose optimality assumptions. That is, we want a reward that results in performance similar to the observed trajectories, but values somewhat above or somewhat below should be equally admissible. Our new polytope for this space of weights is then:
\begin{align}
   \mathcal{P_C} = \left\{ w\,:\begin{array}{rl}
        \dfrac{1}{|1+\varepsilon|} w^T\mu_b \leq  w^T\mu \leq \dfrac{1}{|1-\varepsilon|} w^T\mu_b
        \end{array} \quad \forall \mu \in \mathcal{P_F} \right\},
\label{eqn:p_c}
\end{align}
The parameter $\varepsilon$ that determines the threshold on the consistency polytope is tuned according to our confidence in the batch data; trajectories from severely biased experts may warrant larger $\varepsilon$.

The batch setting also requires computational changes as we do not have access to a simulator to calculate exact feature expectations $\mu$; we must instead estimate them from available data. We do so by adapting off-policy evaluation methods to estimate the representation of a policy in feature space. Specifically, we use per-decision importance sampling (PDIS \citep{precup2000eligibility}) for off-policy value evaluation, to get a consistent, unbiased estimator of $\mu$: $\hat{\mu} = \frac{1}{N}\sum^N_{n=1}{\sum^T_{t=0}\gamma^t{\rho_t^{(n)}} \phi\left(s_t^{(n)}\right)}$,
where importance weights $\rho_t^{(n)} = {\pi(a_t^{n}|s_t^{n})}/{\pi_b(a_t^{n}|s_t^{n})}$. 
Together with the feature expectations of the observed experts (obtained by simple averaging across trajectories), we can now evaluate the constraint in Eq~\ref{eqn:p_c}.
\begin{prop}
The set of weights $\mathcal{P_C}$ defines a closed convex set, given access to exact MDP solver.
\vspace{-4mm}
\begin{proof}
As the redefined constraints are still linear in $w$, that is, of the form $w^TA \leq b$, the convexity argument in \citep{huang2018learning} holds. 
\end{proof}
\end{prop}
\vspace{-3mm}
In Section~\ref{qarp}, we discuss how this assumption of convexity changes given the presence of approximation error in the MDP solver and in estimated feature expectations.
\vspace{-0.05in}
\paragraph{Illustration.} 
We first construct a simple, synthetic task to visualize a polytope of consistent rewards. Consider an agent in a two-dimensional continuous environment, with state defined by position $s_t = [x_t,\, y_t]$ for bounded $x_t$ and $y_t$. At each time $t$, available actions are steps in one of four directions, with random step size $\delta_t \sim \mathcal{N}(0.4, 0.1)$. 
The reward is $r_t = [0.5, 0.5]^T s_t$: the agent's goal is to reach the top-right corner of the 2D map. We use fitted-Q iteration with tree-based approximation \citep{ernst2005tree} to learn a deterministic policy $\pi_b$ that optimizes the reward, then we sample $1000$ trajectories from a {biased} policy (move left with probability $\epsilon$, and $\pi_b$ otherwise) to obtain batch data.

\begin{figure*}[t]
\begin{center}
\vspace{-4mm}
    \centering
     \subfigure{\includegraphics[width=0.28\textwidth]{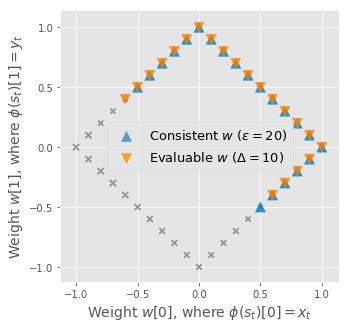} \label{fig:gw1}}
      \subfigure{\includegraphics[width=0.28\textwidth]{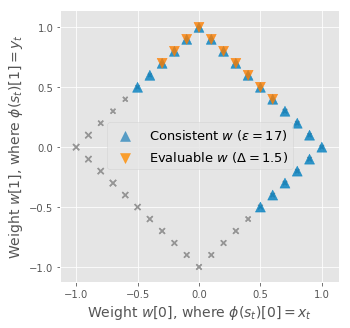} \label{fig:gw2}}
       \subfigure{\includegraphics[width=0.28\textwidth]{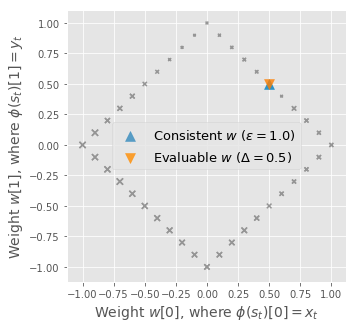} \label{fig:gw4}}
   \vspace{-3mm}
   \caption{Consistent and evaluable polytopes, with different thresholds $\varepsilon$ and $\Delta$ respectively, for a 2D map with true reward $r_t = [0.5, 0.5]^T\phi(s_t)$. Points in the intersection comprise the admissible set.
   }
    \label{fig:gridworld}
\end{center}
\end{figure*}

We then train policies $\pi_w$ optimizing for reward functions $r_t = w^T\phi(s)$ on a set of candidate weights $w \in \mathbb{R}^2$ on the unit $\ell 1$-norm ball. For each policy, a PDIS estimate of the feature expectations $\hat{\mu}_w$ is obtained using the collected batch data. The consistency constraint (Eq~\ref{eqn:p_c}) is then evaluated for each candidate weight vector, with different thresholds $\varepsilon$ (Fig~\ref{fig:gridworld}). For large $\varepsilon\, (\varepsilon \geq 17)$, the set of consistent $w$ includes half of all test weights: given these thresholds, all $w$ for which at least one dimension has a positive weight greater than $0.5$ were determined to yield policies sufficiently close to the batch data, while vectors with large negative weights on either coordinate are rejected. When $\varepsilon$ is reduced to $1.0$, only the reward originally optimized for the batch data, ($w = [0.5, 0.5]$) is admitted by $\mathcal{P_C}$.
\vspace{-0.05in}

\subsection{Evaluable Reward Polytope}
\label{erp}

Our second set of constraints on reward design stem from the need to be able to evaluate and deploy a policy safely in high-risk settings, when further data collection is expensive or infeasible. We interpret this as a condition on confidence in the estimated policy performance: given an estimate for the expected value $\mathbb{E}[\hat{V}^{\pi}] = w^T\hat{\mu}$ of a policy ${\pi}$ and corresponding probabilistic lower bound $V^{\pi}_{lb}$, we constrain the ratio of these values to lie within some threshold $\Delta \geq 0$. A reward function with weights $w$ lies within the polytope of evaluable rewards if $V^{\pi}_{lb} \geq (1-\Delta) w^T\hat{\mu} \,\,\,\forall \mu \in \mathcal{P_F}$, where
$\hat{\mu}$ is our PDIS estimate of feature expectations. 
To formulate this as a linear constraint in the space of reward weights $w$, the value lower bound  $V^{\pi}_{lb}$ must be rewritten in terms of $w$.
This is done by constructing a combination of upper and lower confidence bounds on the policy feature expectations, denoted $\mu^{lb}$. Starting from the empirical Bernstein concentration inequality (Eq~\ref{eqn:thomas-bound}): 

\vspace{-5mm}
{\small{\begin{align} \nonumber
    {V_{lb}} &=\dfrac{1}{N}\smashoperator{\sum\limits_{n=1}^N} \hat{V}_n \!-\! \dfrac{1}{N}\!\sqrt{{\color{Blue}\underbrace{\color{black}\dfrac{\ln(\frac{2}{\delta})}{N\!-\!1}}_{\color{Blue}c_1}} \smashoperator[r]{\sum\limits_{n,n'=1}^N}(\hat{V}_i-\hat{V}_j)^2} \smashoperator{-} {\color{Blue}\underbrace{\color{black}\dfrac{7b\ln(\frac{2}{\delta})}{3(N-1)}}_{\color{Blue}c_2}}
    \\
    &= \dfrac{1}{N}w\!\cdot\!\smashoperator{\sum\limits_{n=1}^N} \hat{\mu}^{(n)} \!-\! sgn(w)\!\cdot\! \dfrac{1}{N}w\!\cdot\!\sqrt{c_1\smashoperator[lr]{\sum\limits_{n,n'=1}^N}\left(\hat{\mu}^{(n)} \!-\! \hat{\mu}^{(n')}\right)^2} \!\smashoperator{-}\! c_2  
    \quad = w^T \hat{\mu}^{lb} - c_2
\label{eqn:factor_lb}
\end{align}}}
\vspace{-4mm}

{where the $k^{th}$ element of $\hat{\mu}^{lb}$---that is, the value of the $k^{th}$ feature that yields the lower bound in the value of the policy---is dependent on the sign of the corresponding element of the weights, $w[k]$:}

\vspace{-4mm}
{\small{
    \begin{align}
    \label{eqn:mu_lb}
     \hat{\mu}^{lb}[k] \!&=\! \begin{cases}
    \dfrac{1}{N}\left[\smashoperator{\sum\limits_{n=1}^N} \hat{\mu}^{(n)} \!\smashoperator{-}\! \sqrt{c_1\smashoperator[lr]{\sum\limits_{n,n'=1}^N}\left(\hat{\mu}^{(n)} - \hat{\mu}^{(n')}\right)^2}\,\right]_k  w[k] \!\geq\! 0 \\
    \dfrac{1}{N}\left[\smashoperator{\sum\limits_{n=1}^N} \hat{\mu}^{(n)} \!\smashoperator{+}\! \sqrt{c_1\smashoperator[lr]{\sum\limits_{n,n'=1}^N}\left(\hat{\mu}^{(n)} - \hat{\mu}^{(n')}\right)^2}\,\right]_k  w[k] \!<\! 0 \\
\end{cases}
\end{align}}} 
\vspace{-3mm}

This definition allows us to incorporate uncertainty in $\hat{\mu}$ when evaluating our confidence in a given policy: a lower bound for our value estimate requires the \emph{lower bound} of $\hat{\mu}$ if the weight is positive, and the \emph{upper bound} if the weight is negative.
Thus, the \emph{evaluable reward polytope} can be written as:
\begin{align}
\label{eqn:poly_e}
\mathcal{P_E} &= \left\{w\,:\,w^T{\mu}^{lb} \leq (1-\Delta) w^T{\mu} \quad \forall \mu \in \mathcal{P_F} \right\} 
\end{align}
The constant $c_2$ in the performance lower bound (Eq~\ref{eqn:factor_lb}) is absorbed by threshold parameter $\Delta$.
\begin{prop}
The set of weights $\mathcal{P_E}$ defines a closed convex set, given access to exact MDP solver.
\vspace{-4mm}
\begin{proof}
As in $\mathcal{P_C}$, constraints on  $\mathcal{P_E}$ are linear in $w$. 
\end{proof}
\end{prop}
\paragraph{Illustration.} In order to visualize an example polytope for evaluable rewards (Eq~\ref{eqn:poly_e}), we return to the two-dimensional map described in Section~\ref{crp}. As before, we begin with a batch of trajectories collected by a biased $\epsilon$-greedy expert policy trained on the true reward. We use these trajectories to obtain PDIS estimates $\hat{\mu}$ for policies trained with a range of reward weights $w$ on the $\ell 1$-norm ball. We then evaluate $\hat{\mu}^{lb}$, and in turn the hyperplanes defining the intersecting half-spaces of the evaluable reward polytope, for each $w$. 
Plotting the set of evaluable reward vectors for different thresholds $\Delta$, we see substantial overlap with the consistent reward polytope in this environment, though neither polytope is a subset of the other (Fig~\ref{fig:gw1}). We also find that in this setting, the value of the evaluability constraint is asymmetric about the true reward---unlike the consistency metric---such that policies trained on penalizing $x_t\, ( w[0] < 0)$, hence favoring movement left, can be evaluated to obtain a tighter lower bound than weights that learn policies with movement down, which is rarely seen in the biased demonstration data (Fig~\ref{fig:gw2}).
Finally, tightening the threshold further to $\Delta=0.5$ (Fig~\ref{fig:gw4}) the set of accepted weights is again just the true reward, as for the consistency polytope.

\subsection{Querying Admissible Reward Polytope}
\label{qarp}
Given our criteria for consistency and evaluability of reward functions, we need a way to access the sets satisfying these constraints. These sets cannot be explicitly described as there are infinite policies with corresponding representations $\mu$, and so infinite possible constraints; instead, we construct a \emph{separation oracle} to access points in this set in polynomial time (Alg~\ref{alg:sep-oracle}). 
A separation oracle tests whether a given point $w'$ lies in polytope of interest $\mathcal{P}$, and if not, outputs a separating hyperplane defining some half-space $w^TA \leq b$, such that $\mathcal{P}$ lies inside this half-space and $w'$ lies outside of it. The separation oracle for the polytope of \emph{admissible} rewards evaluates both consistency and evaluability to determine whether $w'$ lies in the intersection of the two polytopes, which we define as our admissible polytope $\mathcal{P}_{adm}$. If a constraint is not met, it outputs a new hyperplane accordingly.

It should be noted that the RL problems of interest to us are typically large MDPs with continuous state spaces, as in the clinical setting of managing mechanical ventilation in the ICU, and moreover, because we are optimizing policies given only batch data, we know we can only expect to find \emph{approximately} optimal policies. The use of PDIS estimates $\hat{\mu}$ of the true feature expectations in the batch setting introduces an additional source of approximation error. It has been shown that Alg~\ref{alg:sep-oracle} with an approximate MDP solver produces a \emph{weird separation oracle} \citep{huang2018learning}, one that does not necessarily define a convex set.  However, it does accept all points in the queried polytope, and can thus still be used to produce admissible weights.
\setlength{\textfloatsep}{10pt}
  \begin{algorithm}[tb] 
  {\textsc{Input:} $w \in \mathbb{R}^k$, behaviour policy $\mu_b$, threshold parameters $\varepsilon, \Delta$ }
  
  Solve MDP with weights $w$ for optimal policy 
  $\mu=\argmax_{\mu'} w^T\mu' $; evaluate lower bound $\mu^{lb}$ \\
  \uIf{$w^T\mu < \frac{1}{|1+\varepsilon|}w^T\mu_b$}{
  {$w \notin \mathcal{P_C} \Rightarrow \textsc{Reject } w$}; \text{output halfspace} $\{w^T(|1+\varepsilon|\mu - \mu_b) \geq 0 \} $}
  \uElseIf{$w^T\mu > \frac{1}{|1-\varepsilon|}w^T\mu_b$}{
  {$w \notin \mathcal{P_C} \Rightarrow \textsc{Reject } w$}; \text{output halfspace} $\{w^T(|1-\varepsilon|\mu - \mu_b) \leq 0 \} $}
  \uElseIf{$w^T\mu < \frac{1}{(1-\Delta)}w^T\mu^{lb}$}{
  {$w \notin \mathcal{P_E} \Rightarrow \textsc{Reject } w$}; \text{output halfspace} $\{w^T((1-\Delta)\mu - \mu^{lb}) \geq 0 \} $}
  \Else{$w \in \mathcal{P_C} \cap \mathcal{P_E} = \mathcal{P}_{adm} \Rightarrow$ \textsc{Accept} $w$}
   \caption{Separation oracle $SO_{adm}$ for admissible $w$}
   \label{alg:sep-oracle}
\end{algorithm} 
Returning to our 2D map (Fig~\ref{fig:gridworld}), the \emph{admissible reward polytope} is the set of weights accepted by both the consistent and evaluable polytopes. The choice of thresholds $\varepsilon$ and $\Delta$ respectively is important in obtaining a meaningfully restricted, non-empty set to choose rewards from. These thresholds will depend on the extent of exploratory or sub-optimal behaviour in the batch data, and the level of uncertainty acceptable when deploying a new policy. 
\vspace{-0.05in}
\paragraph{Finding the Nearest Admissible Reward} 
\label{fnar}
With a separation oracle $SO_{adm}$ for querying whether a given $w$ lies in the admissible reward polytope, we optimize linear functions over this set using, e.g., the ellipsoid method for exact solutions or---as considered here---the iterative \emph{follow-perturbed-leader} (FPL) algorithm for computationally efficient approximate solutions \citep{kalai2016efficient}. To achieve our goal of aiding reward specification for a designer with existing but imperfectly known goals, we pose our optimization problem as follows (Alg~\ref{alg:fpl}): given initial reward weights $w_0$ proposed by the agent designer, we first test whether $w_0$, with some small perturbation, lies in the admissible polytope $\mathcal{P}_{adm}$, which we define by training a policy $\pi_0$ approximately optimizing this reward. If it does not lie in $\mathcal{P}_{adm}$, we return new weights $w \in \mathcal{P}_{adm}$ that minimize distance $\|w-w_{init}\|_2$ from the proposed weights. This solution is then perturbed and tested in turn. The constrained linear program solved at each iteration scales in constant time with the dimensionality of $w$. Our final reward weights and a randomized policy are the average across the approximate solutions in each iteration. This policy optimizes a reward that is the closest admissible reward to the original goals of the designer. 

\setlength{\textfloatsep}{10pt}
\begin{algorithm}[t]
\SetAlgoLined
\textsc{Input:} Initial weights $w_{0}\in\mathbb{R}^k$, iterations $T$,  $\delta = \frac{1}{k\sqrt{T}}$, $t=0$ 

\While{$t \leq T$}{

\textsc{\tiny 1.} Let $r_t = \sum_{i=1}^{t-1}(w_i + p_t)\cdot \phi(\cdot)$, where $p_t \sim \mathcal{U}[0, \frac{1}{\delta}]^k$; solve for $\pi_t = \argmax_{\pi}V^{\pi}|r_t$

\textsc{\tiny 2.} Let $\mu_t=\mu(\pi_t) + q_t$, where $q_t \sim \mathcal{U}[0, \frac{1}{\delta}]^k$; evaluate constraints defining $\mathcal{P}_{adm}$

\textsc{\tiny 3.} Solve for {$w_t := \argmin_{w\in\mathcal{P}_{adm}}\|w-w_{init}\|_2$}

\textsc{\tiny 4.} Let $t := t+1$ }

\vspace{-3mm}
{\textsc{Output:} $\pi_\textit{final} = \frac{1}{T}\sum\limits_{i=1}^{T}\pi_t$; $\bar{w} = \frac{1}{T}\sum\limits_{i=1}^{T}w_t$}
\caption{\emph{Follow-perturbed-leader} for admissible $w$.}
\label{alg:fpl}
\end{algorithm}

\section{Experiment Design}
\label{exp}

\paragraph{Benchmark Tasks}
\label{bcc}

We illustrate our approach to determining admissible reward functions on three benchmark domains with well-defined objectives: classical control tasks Mountain Car and Acrobot, and a simulation-based treatment task for HIV patients. The control tasks, implemented using OpenAI Gym~\citep{brockman2016openai}, both have a continuous state space and discrete action space, and the objective is to reach a terminal goal state. 
To explore how the constrained polytopes inform reward design for these tasks, an expert behaviour policy is first trained with data collected from an exploratory policy receiving a reward of $-1$ at each time step, and $0$ once the goal state is reached. A batch of 1000 trajectories is collected by following this expert policy with Boltzmann exploration, mimicking a sub-optimal expert. 
Given these trajectories, our task is to choose a reward function that allows us to efficiently learn an optimal policy that is i) consistent with the expert behaviour in the trajectories, and ii) evaluable with acceptably tight lower bounds on performance. We limit the reward function $r_t = w^T\phi(s_t)$ in each task to a weighted sum of three features, $\phi(s)\in \mathbb{R}^3$, chosen to include sufficient information to learn a meaningful policy while allowing for visualization.
For Mountain Car, we use quantile-transformed position, velocity, and an indicator $\pm1$ of whether the goal state has been reached. For Acrobot, $\phi(s)$ comprises the quantile-transformed cosine of the angle of the first link, angular velocity of the link, and an indicator $\pm1$ of whether the goal link height is satisfied. We sweep over weight vectors on the 3D $\ell1$-norm ball, training policies with the corresponding rewards, and filtering for admissible $w$.

The characterization of a good policy is more complex in our third benchmark task, namely treatment recommendation for HIV patients, modeled by a linear dynamical system \citep{ernst2006clinical}. Again, we have a continuous state space and four discrete actions to choose from: no treatment, one of two possible drugs, or both in conjunction. The true reward in this domain is given by: $R = -0.1V + 10^3E - 2\cdot10^4(0.7d_1)^2 - 2\cdot10^3(0.3d_2)^2$, where $V$ is the viral count, $E$ is the count of white blood cells (WBC) targeting the virus, and $d_1$ and $d_2$ are indicators for drugs 1 and 2 respectively. We can rewrite this function as $r = w^T\phi(s)$, where $\phi(s) = [V, c_0 E, c_1d_1 + c_2d_2] \in \mathbb{R}^3$, with constants $c_0, c_1$ and $c_2$ set such that weights $w = [-0.1, 0.5, 0.4]$ reproduce the original function. Again, the low dimensionality of $\phi(s)$ is simply for the sake of interpretability. An expert policy is trained using this true reward, and a set of sub-optimal trajectories are collected by following this policy with Boltzmann exploration. Policies are then trained over weights $w, ||w||_1 = 1$ to determine the set of admissible rewards.
\vspace{-0.05in}
\paragraph{Mechanical Ventilation in ICU}
\vspace{-0.05in}
We use our methods to design rewards for management of invasive mechanical ventilation in critically ill patients \citep{prasad2017reinforcement}. Mechanical ventilation refers to the use of external breathing support to replace spontaneous breathing in patients with compromised lung function. It is one of the most common, as well as most costly, interventions in the ICU~\citep{saunders2018evaluating}. Timely \emph{weaning}, or removal of breathing support, is crucial to minimizing risks of ventilator-associated infection or over-sedation, while avoiding failed breathing tests or reintubation due to premature weaning. Expert opinion varies on how best to trade off these risks, and clinicians tend to err towards conservative estimates of patient wean readiness, resulting in extended ICU stays and inflated costs. 

We look to design a reward function for a weaning policy that penalizes prolonged ventilation, while weighing the relative risks of premature weaning such that the optimal policy does not recommend strategies starkly different from clinician behaviour, and the policies can be evaluated for acceptably robust bounds on performance using existing trajectories.
We train and test our policies on data filtered from the MIMIC III data \citep{johnson2016mimic} with 6,883 ICU admissions from successfully discharged patients following mechanical ventilation, preprocessed and resampled in hourly intervals. 
The MDP for this task is adapted from \citep{prasad2017reinforcement}: the patient state $s_t$ at time $t$ is a 32-dimensional vector that includes demographic data, ventilator settings, and relevant vitals. We learn a policy with binary action space $a_t \in [0, 1]$, for keeping the patient off or on the ventilator, respectively.
The reward function $r_t = w^T\phi(s_t, a_t)$ with $\phi(s,a) \in \mathbb{R}^3$ includes (i) a penalty for more than 48 hours on the ventilator, (ii) a penalty for reintubation due to unsuccessful weaning, and (iii) a penalty on physiological instability when the patient is \emph{off} the ventilator based on abnormal vitals. Our goal is to learn the relative weights of these feedback signals to produce a consistent, evaluable reward function.

\section{Results and Discussion}
\subsection{Benchmark Control Tasks}
\begin{table*}[t]
\captionsetup{justification=centering}
\caption{Analysing admitted weights $w$ for each of the three benchmark tasks. Polytope thresholds are set by choosing a small $\Delta$ and a large $\varepsilon$ for an admissible set of size 3.}
\label{table:admissible_stas}
\begin{center}
\begin{small}
\begin{sc}
\begin{tabular}{l | c c c}
\toprule
Task & Top 3 Admissible weights & $\varepsilon$ ($\mathcal{P_C}$) & $\Delta$ ($\mathcal{P_E}$) \\
\midrule
Mountain Car & $[0.0, 0.0, 1.0]^T,\, [0.2, 0.0, 0.8]^T,\, [0.2, -0.6, 0.2]^T$ & 0.98 & 0.50 \\
Acrobot & $[-0.6, -0.2, 0.2]^T,\, [-0.6, 0.0, 0.4]^T,\, [-0.8, 0.0, 0.2]^T$ & 1.00 & 1.80 \\
HIV Simulator& $[0.0, 1.0, 0.0]^T,\, [0.4,  0.6, 0.0]^T,\, [0.2, 0.8, 0.0 ]^T$ & 1.00 & 0.60 \\
\bottomrule
\end{tabular}
\end{sc}
\end{small}
\end{center}
\vskip -0.1in
\end{table*}
\begin{figure}
\vskip -0.1in
    \centering
    \includegraphics[width=0.95\textwidth]{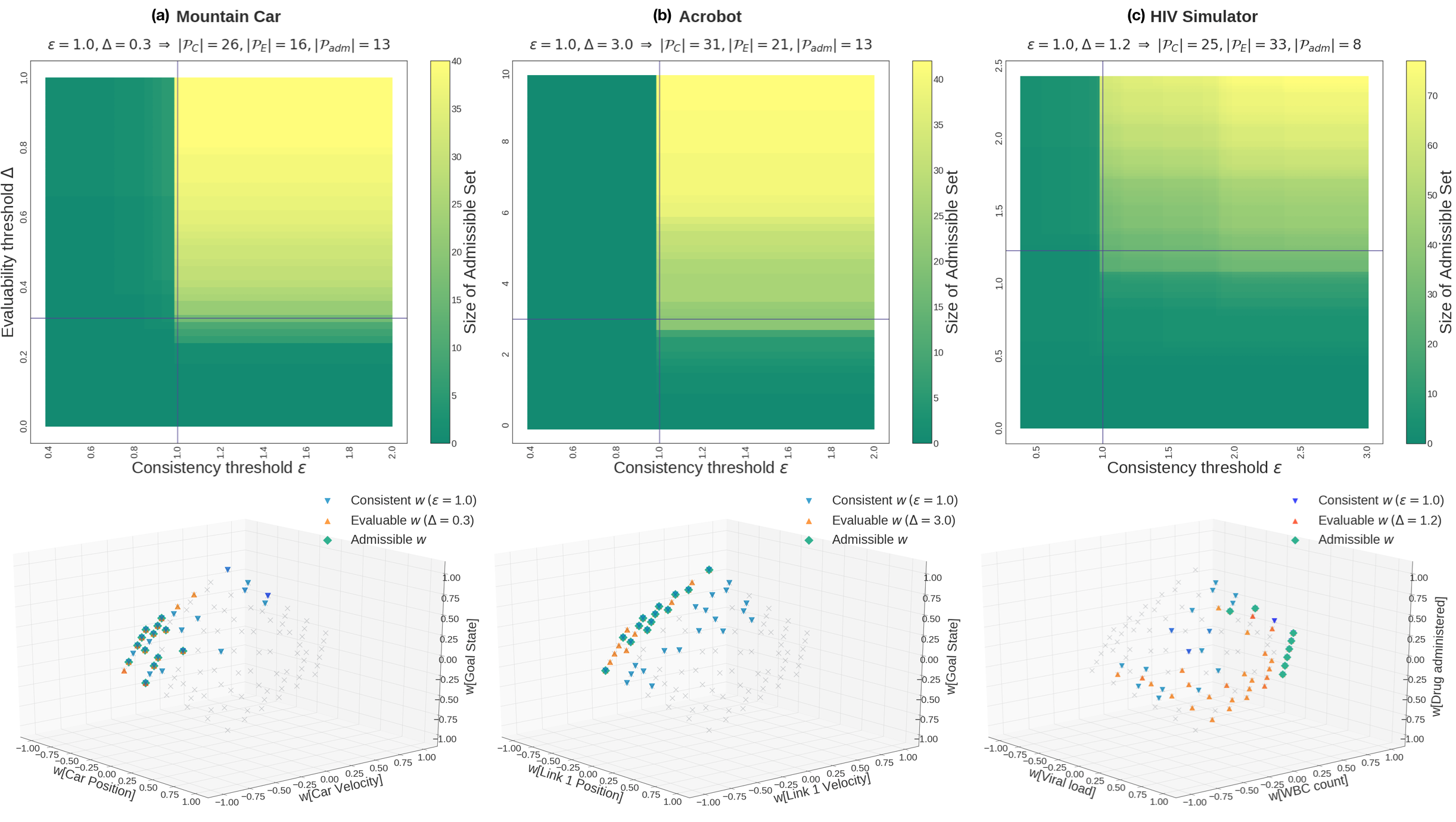}
     \caption{Admissible polytope size for varying thresholds on consistency ($\varepsilon$) and evaluability ($\Delta$), and distribution of admitted weights for fixed $\varepsilon, \Delta$, in: (a) Mountain Car (b) Acrobot (c) HIV Simulator. Note that admitted rewards for each task typically correspond to positive weights on the goal state.}
     \label{fig:all}
\end{figure}

\paragraph{Admissible $\bm{w}$ are clustered near true rewards.} We analyze reward functions from the sweep over weight vectors on the $\ell1$-norm unit ball for each benchmark task (Section~\ref{bcc}) by first visualizing how the space of weights accepted by the consistency and evaluability polytopes---and therefore the space $\mathcal{P}_{adm}$ at the intersection of these polytopes---changes with the values of thresholds $\varepsilon$ and $\Delta$. Alongside this, we plot the set of admitted weights produced by arbitrarily chosen  thresholds (Fig~\ref{fig:all}).
In all three tasks, we find that the admitted weights form distinct clusters; these are typically at positive weights on goal states in the classic control tasks, and at positive weights on WBC count for the HIV simulator, in keeping with the rewards optimized by the batch data in each case. We could therefore use this naive sweep over weights to choose a vector within the admitted cluster that is closest to our initial proposed function, or to our overall objective. For instance, if in the HIV task we want a policy that prioritizes minimization of side effects from administered drugs, we can choose specifically from admissible rewards with negative weight on the treatment term.
\vspace{-0.05in}
\paragraph{Analysis of admissible $\bm{w}$ can lend insight into reward shaping for faster policy learning.} We may wish to shortlist candidate weights by setting more stringent thresholds for admissibility. 
We mimic this design process as follows: prioritizing evaluability in each of our benchmark environments, we choose the smallest possible $\Delta$ and large $\varepsilon$ for an admissible set of exactly three weights (Table~\ref{table:admissible_stas}). This reflects a typical batch setting, in which we want high-confidence performance guarantees; we also want to allow our policy to deviate when necessary from the available sub-optimal expert trajectories. For Mountain Car, our results show that two of the three vectors assign large positive weights to reaching the goal state; all assign zero or positive weight to the position of the car. The third, $w=[0.2, -0.6, 0.2]$ is dominated by a large negative weight on velocity; this may be interpreted as a kind of reward shaping: the agent is encouraged to first move in reverse to achieve a negative velocity, as is necessary to solve the problem.
The top three vectors for Acrobot also place positive weights on the goal state, but large negative weights on the position of the first link. Again, this likely plays a shaping role in policy optimization by rewarding link displacement.
\vspace{-0.05in}
\paragraph{FPL can be used to correct biased reward specification in the direction of true reward.} We use the HIV treatment task to explore how iterative solutions for admissible reward (Alg \ref{alg:fpl}) can improve a partial or flawed reward specified by a designer. For instance, a simple first attempt by the designer at a reward function may place equal weights on each component of $\phi(s)$, with the polarity of weights---whether each component should elicit positive feedback or incur a penalty---decided by the designer's domain knowledge; here, the designer may suggest $w_0 = \frac{1}{3}[-1, 1, -1]^T$. We run Alg~\ref{alg:fpl} for twenty iterations with this initial vector and thresholds $\varepsilon=1.0, \Delta = 0.6$ and average over the weights from each iteration. This yields weights $\bar{w} = [-0.11, 0.57, -0.32]^T$, redistributed to be closer to the reward function being optimized in the batch data. This pattern is observed with more extreme initial rewards functions too; if e.g., the reward proposed depends solely on WBC count, $w_0 = [0, 1, 0]$, then we obtain weights $\bar{w} = [-0.14, 0.83, -0.04]$ after twenty iterations of this algorithm such that appropriate penalties are introduced on viral load and administered drugs.

\begin{figure}[t]
\begin{floatrow}
\ffigbox[5.9cm]{%
  \includegraphics[width=0.35\textwidth]{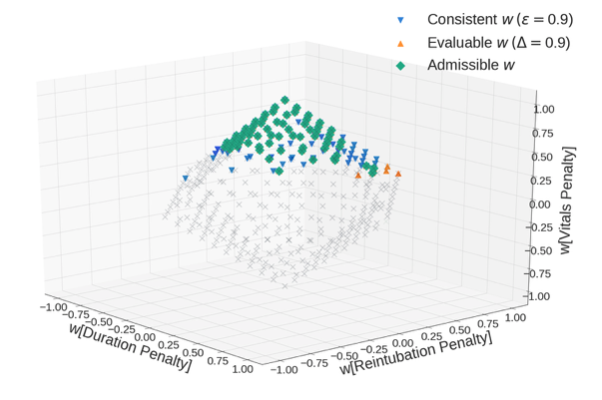}%
}{%
 \caption{{VentICU: Admissible polytope}}
 \label{fig:ventadm}
}
\capbtabbox{%
\begin{sc}
\begin{small}
\renewcommand{\arraystretch}{1.2}
\begin{tabular}{r | c | c | c}
        \toprule
        Initial $w$ & $N_\textit{eff}$ & Final $w$ & $N_\textit{eff}$ \\
        \midrule
        $[1. , 0. , 0.]$ & 2.0 & $[0.82, 0.13, 0.05]$ & 13.3 \\
        $[0., 1., 0.]$ & 24.6 & $[\text{-}0.07, 0.82, 0.08]$ & 32.7 \\
        $[ 0.,  0., 1.]$ & 113.1 & $[ 0.25, \text{-}0.06, 0.68]$ & 119.1 \\
        $\frac{1}{3}[1.,  1., 1.]$ & 79.9 & $[0.16,0.43,0.4]$ & 95.6 \\
        \bottomrule
        \end{tabular}
\end{small}
\end{sc}
}{%
  \caption{VentICU: Effective sample sizes}%
  \label{table:ventess}
}
\end{floatrow}
\end{figure}
\subsection{Mechanical Ventilation in ICU (VentICU)}
\paragraph{Admissible $\bm{w}$ may highlight bias in expert behaviour.} We apply our methods to choose a reward function for a ventilator weaning policy in the ICU, given that we have access only to historical ICU trajectories with which to train and validate our policies. When visualizing the admissible set, with $\varepsilon=\Delta=0.9$, we find substantial intersection in the consistent and evaluable polytopes (Fig~\ref{fig:ventadm})). Admitted weights are clustered at large weights on the penalty term for physiological instability while off the ventilator, favouring policies that are conservative in weaning patients. We can tether a naive reward that penalizes duration on the ventilator, $w = [1, 0, 0]$ to the space of rewards that are consistent with this conservative behaviour. Using FPL to search for a reward within the admissible set given this initial vector yields $\bar{w} = [0.82, 0.13, 0.05]$, with non-zero penalties on reintubation and physiological instability when off ventilation. 
\vspace{-0.05in}
\paragraph{FPL improves effective sample size for learnt policies.} To verify whether weights from the admissible polytope enable higher confidence policy evaluation, we explore a simple proxy for variance of an importance sampling-based estimate of performance: the effective sample size $N_\textit{eff}$ = $(\sum^N_n\rho_n)^2/\sum^N_n\rho_n^2$ of the batch data~\citep{wiegand1968kish}, where $\rho_n$ is importance weight of trajectory $n$ for a given policy. Testing a number of naive initializations of $w$, we find that effective sample size is consistently higher for weights following FPL (Table~\ref{table:ventess}). This indicates that the final weights induce an optimal policy that is better represented in the batch data than the policy from the original weights.

\section{Conclusion}

In this work, we present a method for reward design in RL using batch data collected from sub-optimal experts. We do this by constraining rewards to those yielding policies within some distance of the policies of domain experts; the policies inferred from the admissible rewards also provide reasonable bounds on performance. Our experiments show how rewards can be chosen in practice from the space of functions satisfying these constraints, and illustrate this on the problem of weaning clinical patients from mechanical ventilation.
Effective reward design for RL in safety-critical settings is necessarily an iterative process of deployment and evaluation, to push the space of observed behaviour incrementally towards policies consistent and evaluable with respect to our ideal reward.

\bibliographystyle{local}
\bibliography{ref}

\newpage
\appendix
\thispagestyle{empty} 
\onecolumn
\section{Defining Admissible Rewards for High Confidence Off-Policy Evaluation:
Supplementary Material}

\setcounter{section}{1}
\subsection{{Mechanical Ventilation in the ICU: Experimental Details}}
{\textsc{MDP Formulation}}

In learning a policy for weaning patients from mechanical ventilation in the ICU, we model patients as an MDP; at each time $t$, the state of the patient $s_t$ is modelled by a 32-dimensional feature vector, adapting from \cite{prasad2017reinforcement}. These state features are listed in the table below.

\begin{table*}[th]
\renewcommand{\arraystretch}{1.2}
\begin{center}
\begin{small}
\begin{tabular}{ r | l }
\toprule
 & \sc{State Features} \\
\midrule
\sc{Demographics} &  Age, Gender, Ethnicity, Admission Weight, First ICU Unit \\
\hline
\sc{Ventilator Settings} & Ventilator mode, Inspired $O_2$ fraction ($FiO_2$), PEEP set, $O_2$ Flow  \\
\hline
\sc{Measured Vitals} & Heart Rate, Respiratory Rate, $O_2$ saturation pulseoxymetry, \\ & 
Non Invasive Blood Pressure mean, Non Invasive Blood Pressure systolic, \\ &
Non Invasive Blood Pressure diastolic, Mean Airway Pressure, Tidal Volume, \\ &
Arterial pH, Richmond-RAS Scale, Peak Insp. Pressure, 
Plateau Pressure, \\ & Arterial $CO_2$ Pressure, Arterial $O_2$ pressure  \\
\hline
\sc{Input Sedation} & Propofol, Fentanyl, Midazolam, Dexmedetomidine, Morphine Sulfate, \\ & Hydromorphone, Lorazepam \\
\hline
\sc{Other} & Consecutive duration into ventilation $(D)$, Number of reintubations $(N)$ \\
\bottomrule
\end{tabular}
\end{small}
\end{center}
\end{table*}
\vspace{-3mm}

The MDP action space is simply a binary choice between maintaining/placing patients on mechanical ventilation in the next time step ($a_t = 1$), or maintaining/weaning patients off the ventilator ($a_t = 0$). 

The reward function for the experiments in this paper is modeled as $r_t = w^T\phi(\cdot)$, where $\phi(\cdot) \in \mathbb{R}^3$ is given by:

\renewcommand*{\arraystretch}{1.1}
$\phi = \begin{bmatrix}
   -\min(0, \tanh{0.1(D_t - 48)}) \cdot \mathbbm{1}[{a_t = 1}]  \\
   -\mathbbm{1}[\exists t'>t \text{ such that } N_{t'} > N_t) \cdot  \mathbbm{1}[{a_t = 0}] \\
     - \frac{1}{|V|}\sum_v^V{(v < v_\text{min} || v < v_\text{max}) } \cdot \mathbbm{1}[{a_t = 0}]
\end{bmatrix}$ 

where $D_t$ is duration into ventilation at time $t$ in an admission, $N_t$ is the number of reintubations, $v \in V$ are physiological parameters each with  normal range $[v_\text{min}, v_\text{max}]$, and $V = \{\text{Ventilator settings, Measured vitals}\}$. The three terms in $\phi(\cdot)$ represent penalties on duration of ventilation, reintubation, and abnormal vitals, respectively.

\textsc{Learning Optimal Policy}

In order to learn an optimal policy for this MDP, we run Fitted Q-iteration (FQI, \cite{ernst2005tree}) with function approximation using extremely randomized trees. We partition our dataset of 6,883 admissions into 3,000 training episodes and 3,883 test episodes, and run FQI over 100 iterations on the training set, with discount factor $\gamma=0.9$. We then use the approximated Q-function to learn a binary policy, again using extremely randomized trees. 

\textsc{Calculating effective sample size $N_\textit{eff}$}

In order to evaluate the Kish effective sample size $N_\textit{eff}$ for a given policy, we subsample admissions in our test data to obtain trajectories of approximately 20 timesteps in length, and calculate importance weights $\rho_n$ for the policy considered using these subsampled trajectories.

\newpage
\subsection{Follow-Perturbed-Leader: Scaling Dimensionality of $\bf{\phi(\cdot)}$}

Although we only present results with $\phi$ dimensionality $\leq 3$, for the sake of visualization, the iterative algorithm presented can be scaled to much larger $\phi$, as the complexity of the linear program solved at each iteration is dependent only on the number of constraints, not the number of dimensions we solve for. We verify this by running FPL for various $\phi$ (with additional features assigned Gaussian noise) and noting that each iteration completes in constant time.

\begin{table}[th]
\renewcommand{\arraystretch}{1.}
\begin{center}
\begin{small}
\begin{tabular}{ c | c }
\toprule
$dim(\phi(\cdot))$ & Time per FPL iteration $(s)$ \\
\midrule
3 & 41.6890809536 \\
5 & 42.4180600643\\
7 & 42.2197260857\\
10 & 42.9907691479\\
30 & 42.3232078552\\
100 & 42.6504528522\\
\bottomrule
\end{tabular}
\end{small}
\end{center}
\vskip 0.1in
\end{table}
\thispagestyle{empty} 

\end{document}